\pdfoutput=1

\documentclass[11pt]{article}

\usepackage[preprint]{acl}

\usepackage{times}
\usepackage{latexsym}

\usepackage[T1]{fontenc}

\usepackage[utf8]{inputenc}

\usepackage{microtype}

\usepackage{inconsolata}

\usepackage{colortbl}
\definecolor{mygray}{gray}{.9}
\definecolor{mypink}{rgb}{.99,.91,.95}
\definecolor{mycyan}{cmyk}{.3,0,0,0}

\usepackage{tikz}
\usepackage[edges]{forest}

\usepackage{graphicx}

\title{Privacy in LLM-based Recommendation: Recent Advances and Future Directions}

\author{Sichun Luo$^1$, Wei Shao$^1$, Yuxuan Yao$^1$, Jian Xu$^2$, Mingyang Liu$^1$, Qintong Li$^3$, \\\textbf{Bowei He$^1$, Maolin Wang$^1$, Guanzhi Deng$^1$, Hanxu Hou$^4$, Xinyi Zhang$^5$, Linqi Song$^1$} \\
  $^1$CityUHK \quad $^2$Tsinghua University \quad $^3$HKU \quad $^4$Dongguan University of Technology \quad $^5$CUEB\\
  \texttt{sichun.luo@my.cityu.edu.hk, linqi.song@cityu.edu.hk} 
  }

\begin{document}
\maketitle
\begin{abstract}
Nowadays, 
large language models (LLMs) have been integrated with conventional recommendation models to improve recommendation performance.
However, while most of the existing works have focused on improving the model performance, the privacy issue has only received comparatively less attention. In this paper, we review recent advancements in privacy within LLM-based recommendation, categorizing them into privacy attacks and protection mechanisms. Additionally, we highlight several challenges and propose future directions for the community to address these critical problems.
\end{abstract}

\section{Introduction}

Recommender systems have been proposed to address the information overload problem \cite{luo2022personalized,luo2022hysage,luo2023improving,10.1145/3664927}. Recently, large language models (LLMs) have shown great ability in natural language understanding, generation, and reasoning \cite{zhao2023survey,luo2024language,wang2023mathcoder,yao2024can}.
They have also been integrated into recommendation systems, achieving significant progress and reshaping the paradigms of recommendation \cite{bao2023tallrec,lin2023can,zhao2024recommender}.

Despite the extensive research on LLM-based recommendation systems, most studies have centered on model accuracy, with less attention dedicated to privacy issues. However, safeguarding privacy is paramount due to the critical nature of private information. 
Thus privacy issues are not peripheral; they are central to the ethical use of these technologies and warrant thorough investigation. 

Several surveys have explored LLM-based recommendation systems \cite{lin2023can,li2023large,zhao2024recommender}, privacy issues in recommendation systems \cite{huang2019privacy,fan2022comprehensive,ogunseyi2023systematic}, and privacy concerns specific to LLMs \cite{pan2020privacy,neel2023privacy,yao2024survey,yan2024protecting}. Nevertheless, there has been no comprehensive study that specifically focus on privacy in LLM-based recommendation systems, leaving a significant gap in the literature.

In this paper, we aim to bridge this gap. We first provide an overview of LLM-based recommendation systems and examine the privacy issues inherent to LLMs and their applications in recommendations. Thereafter, we discuss various privacy attacks and the corresponding privacy protection mechanisms. 
Subsequently, we point out some challenges and future directions.
Our contributions are threefold:

\begin{itemize}
    \item 
    To the best of our knowledge, we are the first to conduct an extensive review of recent advances in privacy issues in LLM-based recommendation systems. 
    \item 
    We systematically introduce privacy in LLM-based recommendations, including privacy definition, attacks, and protection methods.
    \item 
    We highlight current challenges and promising research directions for developing privacy-preserving LLM-based recommender systems.
\end{itemize}

\section{Preliminary}

\subsection{LLM-based Recommendation}

LLM-based recommendation can be classified into several categories. Some works use LLMs as feature encoders \cite{harte2023leveraging, rajput2024recommender}, feature engineering tools \cite{wei2024llmrec, luo2024integrating}, scoring/ranking functions \cite{luo2023recranker, lin2023clickprompt}, and for modeling user interaction \cite{liu2023conversational, leszczynski2023talk}. 
Noteworthy advancements have been made in utilizing LLMs as recommenders, either through in context learning \cite{luo2024large,liu2023chatgpt} or supervised fine-tuning \cite{bao2023tallrec,zhang2023recommendation}.
However, while a significant emphasis is placed on model accuracy, the aspect of privacy remains relatively under-explored, thereby presenting a substantial research gap.

\subsection{Privacy Issues}

\subsubsection{Privacy Issues in LLMs}

Privacy concerns pervade the life cycle of LLMs, including pre-training, fine-tuning, and inference stages. We briefly discuss them as follows.

\paragraph{Pre-training Stage}
LLMs are trained on massive datasets, often sourced from publicly available internet data, which can contain sensitive information such as addresses, emails, and Social Security numbers. As model parameters increase, the risk of memorizing and inadvertently outputting this private information also rises, leading to potential privacy breaches.

\paragraph{Fine-tuning Stage}
Fine-tuning (FT) adapts pre-trained LLMs to specific tasks using smaller, domain-specific datasets from fields like healthcare, law, and finance. These datasets can include sensitive information, such as patient privacy data or corporate trade secrets. Despite regulations, LLMs may still memorize private content, and attacks like Membership Inference and Model Inversion can exploit this to reveal sensitive information.

\paragraph{Inference Stage}
During inference, LLMs generate content based on user inputs, which may contain sensitive or personally identifiable information, leading to privacy leaks. Even with assurances from LLM service providers, complete protection of user data is challenging. LLMs can infer user attributes and preferences from interaction history, even without explicit private information, posing significant privacy concerns.

\subsubsection{Privacy in Recommendation}
According to previous works \cite{huang2019privacy,ge2022survey}, the definition of privacy shares similar ingredients.
This includes sensitive user attributes such as identity, gender, age, and address, which necessitate stringent access restrictions. The concept of ownership dictates that only authorized entities—whether individual users or the platform itself—are permitted to access and control private information. However, these systems face significant threats from malicious entities, both internal and external, who seek to gain unauthorized access or manipulate this information. These adversaries often exploit auxiliary public information to facilitate their infiltration or attacks. Consequently, the primary goal of privacy protection within recommender systems is to uphold the ownership of private information and develop effective countermeasures against such threats. This ensures that users' sensitive data remains secure and their privacy intact, fostering trust in the technology.

\subsubsection{Privacy Issues in LLM-based Recommendation}

Privacy in LLM-based recommendation is the intersection of privacy in LLM and privacy in recommendation, since it may involve the incorporation of these two kinds of models.
For example,
when utilizing LLMs as recommenders, privacy concerns primarily center around the privacy of the LLM itself. Conversely, when employing LLMs as feature augmentors in conjunction with conventional recommender systems, privacy issues are more likely to arise within the recommendation process.


\section{Privacy Attack in LLM-based Recommendation}



Privacy attacks on recommendation include several categories, such as Membership Inference Attacks \cite{zhang2021membership} (i.e., identify whether the target user is utilized to train the target recommendation model),  Property Inference Attacks \cite{ganju2018property} (i.e., infer global sensitive information in the training set), Reconstruction Attacks \cite{meng2019towards} (i.e., using the user’s publicly available attributes and multiple queries to reconstruct their sensitive information), and Model Extraction Attacks \cite{yue2021black} (i.e., extract useful information from model).

There are several kinds of privacy attacks on LLM, such as 
Prompt Hacking \cite{shen2023anything,liu2023prompt}
(i.e. strategically designing and manipulating input prompts so that they can influence LLMs’s output),
Adversarial Attack \cite{zhang2021trojaning,nguyen2024backdoor}
(i.e. manipulating input data to cause the LLM to produce incorrect or unintended outputs),
Gradient Leakage Attack \cite{deng2021tag,balunovic2022lamp}
(i.e. access or infer the gradients or the gradient information, which may obtain access to the model or even compromise its privacy and safety).

Some pioneer works focus on privacy attacks in LLM-based recommendation \cite{zhang2024stealthy,zhang2024lorec}.
\citet{zhang2024stealthy}
reveals that the LLM-based recommendation introduces security vulnerabilities, allowing attackers to subtly manipulate item recommendations without detection,
and further propose a stealthy attack via prompt injection.
Moreover,
\citet{zhang2024lorec} propose
LoRec, a framework that effectively utilizes LLMs to detect and defend against poisoning attacks in sequential recommender systems, significantly improving their robustness.

\section{Privacy Protection in LLM-based Recommendation}
\label{sec:pri_pro}


\subsection{LLM-based Recommendation Unlearning}
\label{sec:mac_unl}

Machine unlearning \cite{bourtoule2021machine} is a privacy-preserving technique aiming at addressing the \textit{"right to be forgotten"} challenge underscored by various privacy regulations \cite{mantelero2013eu,illman2019california}.  Its primary objective is to facilitate the effective and efficient erasure or exclusion of sensitive or unwanted information from the trained machine learning model while maintaining the model performance \cite{nguyen2022survey}. 

Recommendation unlearning, a subset of machine unlearning, is specifically tailored for recommendation scenarios \cite{chen2022recommendation,li2023selective,li2024making}. 
LLM-based recommendation unlearning focuses on the LLM-based recommendation siutation.
E2URec \cite{wang2024towards} enhances unlearning efficiency by updating select LoRA parameters and employs a teacher-student framework to improve the unlearning process.
Besides,
APA \cite{hu2024exact} utilize unique adapters for segmented training data shards, retraining only those adapters impacted by unusable data, and employing parameter-level adapter aggregation with sample-adaptive attention for each test sample.
These methods enable effective and efficient unlearning while preserving recommendation performance.

\subsection{LLM-based Federated Recommendation}
\label{sec:fl}

Recently, some pioneering studies have already equipped the LLM-based recommendation with the paradigm of federated learning to leverage the foundation models while preserving the users' data privacy \cite{zeng2024GPTFedRec,zhang2024FedPA,zhao2024llm4fedrec}. For example, \citet{zhao2024llm4fedrec} adopts a federated split learning manner to selectively retain certain sensitive LLM layers on the client side while offloading non-sensitive layers to the server to reduce the local training burden. Meanwhile, a dynamic balance strategy is also proposed to ensure relatively balanced performance across clients. \citet{zeng2024GPTFedRec} proposes a novel hybrid Retrieval Augmented Generation (RAG) mechanism for extracting the ID-based user patterns and text-based item features, which are then converted into text prompts and fed into the ChatGPT for re-ranking. In this way, generalized features could be extracted from the sparse and heterogeneous data, and the rich knowledge within the pre-trained LLM could be exploited. \citet{zhang2024FedPA} proposes a novel adapter-based foundation model customization for recommendation in a federated setting, where each client trains its local adapter using the private data and the local adapters are further clustered into multiple groups in the server. Finally, the personalized adapter is obtained by combining both the user-level and user-group-level adapters. Then, the pre-trained foundation models equipped with personalized adapters can provide fine-grained recommendation service efficiently.

\subsection{Other Privacy Preserving Techniques}

There are other privacy-preserving techniques have been employed within the broader domain of recommender systems, including non-cryptography-based techniques (e.g., anonymization \cite{luo2013distributed,chang2010towards,sakuma2017recommendation}, randomization \cite{polat2005svd,li2006t}, and differential privacy \cite{berlioz2015applying,shin2018privacy,saleem2021parking}) 
and a myriad of cryptography-based techniques (e.g., homomorphic encryption \cite{erkin2012generating,kim2018efficient,jumonji2021privacy}, secure multiparty computation \cite{shmueli2017secure,kaur2019multi,shmueli2020mediated}, and secret sharing \cite{chen2020secure,chen2021secrec,lin2022generic}). These methods have proven effective in their respective applications, providing robust privacy safeguards. 
Specifically,
RAH \cite{shu2023rah} consists of LLM-based agents and emphasizes user alignment, to address challenges in recommender systems such as accuracy, biases, privacy, and cold-start problems.
\citet{carranza2023privacy} addresses differential privacy in training deep retrieval systems by using DP language models to generate private synthetic queries, improving retrieval quality while maintaining privacy guarantees.


\section{Challenges and Future Directions}

We detail the existing works for privacy in LLM-based recommendation. However, there are also some other potential directions to be explored. In this section, we will illustrate some promising directions for further research on this topic.

\begin{itemize}
    \item

\noindent  \textit{Challenge 1: Lack of Universally Applicability in Privacy-Preserving LLM-based Recommendations.}
The application of LLM-based recommender systems spans a wide array of scenarios, each with unique requirements and specificities. Current research often targets narrow, specific use cases, limiting the broader applicability and adaptability of these methods. This fragmentation hinders the development of universally and comprehensively applicable solutions that can seamlessly integrate across varying contexts and industries.

\textit{Future Directions:}
To address this challenge, there is a pressing need for a unified, general, and comprehensive framework that can accommodate the diverse requirements of different application scenarios while ensuring privacy preservation. This necessitates the development of multi-task, multi-domain solutions that are robust and versatile enough to handle the complexity and variability inherent in real-world applications.

\item 

\noindent \textit{Challenge 2: Efficiency and Effectiveness in Privacy-Preserving LLM-based Recommendations.}
The substantial size of LLMs poses significant challenges in terms of computational efficiency. The sheer scale of these models can lead to resource-intensive operations, making them impractical for many applications where quick, reliable recommendations are crucial. Furthermore, privacy-preserving techniques, such as encryption, often incur additional computational costs. Besides, the trade-off between privacy and accuracy means that some privacy-preserving methods may compromise the model’s effectiveness. Consequently, preserving user privacy effectively and efficiently remains a challenging task.

\textit{Future Directions:}
Innovative approaches are required to enhance the efficiency and effectiveness of LLM-based recommendation systems. Research should focus on optimizing model architectures, leveraging techniques like model pruning \cite{ma2023llm}, quantization \cite{lin2023awq}, and distillation \cite{hsieh2023distilling} to reduce computational overhead without compromising performance. Additionally, exploring methods to improve inference speed and energy efficiency will be crucial for the practical deployment of LLMs in recommender systems.

\item 

\noindent  \textit{Challenge 3: Privacy-Preserving Cloud-Edge Collaboration for LLM-based Recommendation.} Due to the large size of LLM, applying cloud-edge collaborative recommender systems could be a promising direction to relieve the computation and storage burden for edge devices. However, it may not be practical to fine-tune the cloud-side large models without the disclosure of local sensitive data, thus reducing the recommendation performance. Moreover, the tremendous volume of edge users also makes it infeasible to customize the cloud-side model for each individual user. Besides, latency caused by intermediate information transmission and extensive computation overhead highly impacts the user experience.

\textit{Future Directions:} Model compression techniques for LLMs could be a promising direction for deploying recommendation models in the edge devices while parameter-efficient fine-tuning methods \cite{ding2023parameter,han2024parameter}, such as adapter \cite{hu2023llm} and LoRA \cite{hu2021lora}, could also be applied for local model customization, thus avoiding uploading the local data. Considering heterogeneous device capabilities, novel model architectures with multiple exits could also offer a flexible trade-off between the performance and efficiency during the model inference.

\end{itemize}

\section{Conclusion}

In this paper, we provide a comprehensive overview of recent advancements for privacy in LLM-based recommendation. More specifically, we introduce the basic concepts and taxonomy to facilitate a better understanding of this topic. We also summarize the representative methods addressing privacy concerns in LLM-based recommendation systems, covering both privacy attacks and privacy protection. Finally, we analyze potential challenges and suggest possible future research directions in the realm of privacy for LLM-based recommendations.

\bibliography{custom}

\appendix



\end{document}